 \pdfoutput=1
 
 \documentclass[11pt]{article}
 
 \usepackage{acl}
 
 \usepackage{times}
 \usepackage{latexsym}
 \usepackage{booktabs}
 \usepackage{multirow}
 \usepackage{rotating}
 \usepackage[T1]{fontenc}
 
 \usepackage[utf8]{inputenc}
 
 \usepackage{microtype}
 \usepackage{arydshln}
 
 \title{Controllable Mixed-Initiative Dialogue Generation through Prompting}
 
 \author{Maximillian Chen,
     Xiao Yu,
     Weiyan Shi,
     Urvi Awasthi,
     Zhou Yu \\
   Columbia University \\
   \texttt{maxchen@cs.columbia.edu} \\
   \texttt{\{xy2437, ws2634, urvi.awasthi, zy2461\}@columbia.edu}
 }
 
\begin{document}
 \maketitle
 \begin{abstract}
 Mixed-initiative dialogue tasks involve repeated exchanges of information and conversational control. Conversational agents gain control by generating responses that follow particular dialogue intents or strategies, prescribed by a policy planner. The standard approach has been fine-tuning pre-trained language models to perform generation conditioned on these intents. However, these supervised generation models are limited by the cost and quality of data annotation.
 We instead prompt large language models as a drop-in replacement to fine-tuning on conditional generation. We formalize prompt construction for controllable mixed-initiative dialogue. Our findings show improvements over fine-tuning and ground truth responses according to human evaluation and automatic metrics for two tasks: PersuasionForGood and Emotional Support Conversations.
 \end{abstract}
 \section{Introduction}
 \begin{figure}[t]
     \centering
     \scalebox{0.8}{\includegraphics[width=\linewidth]{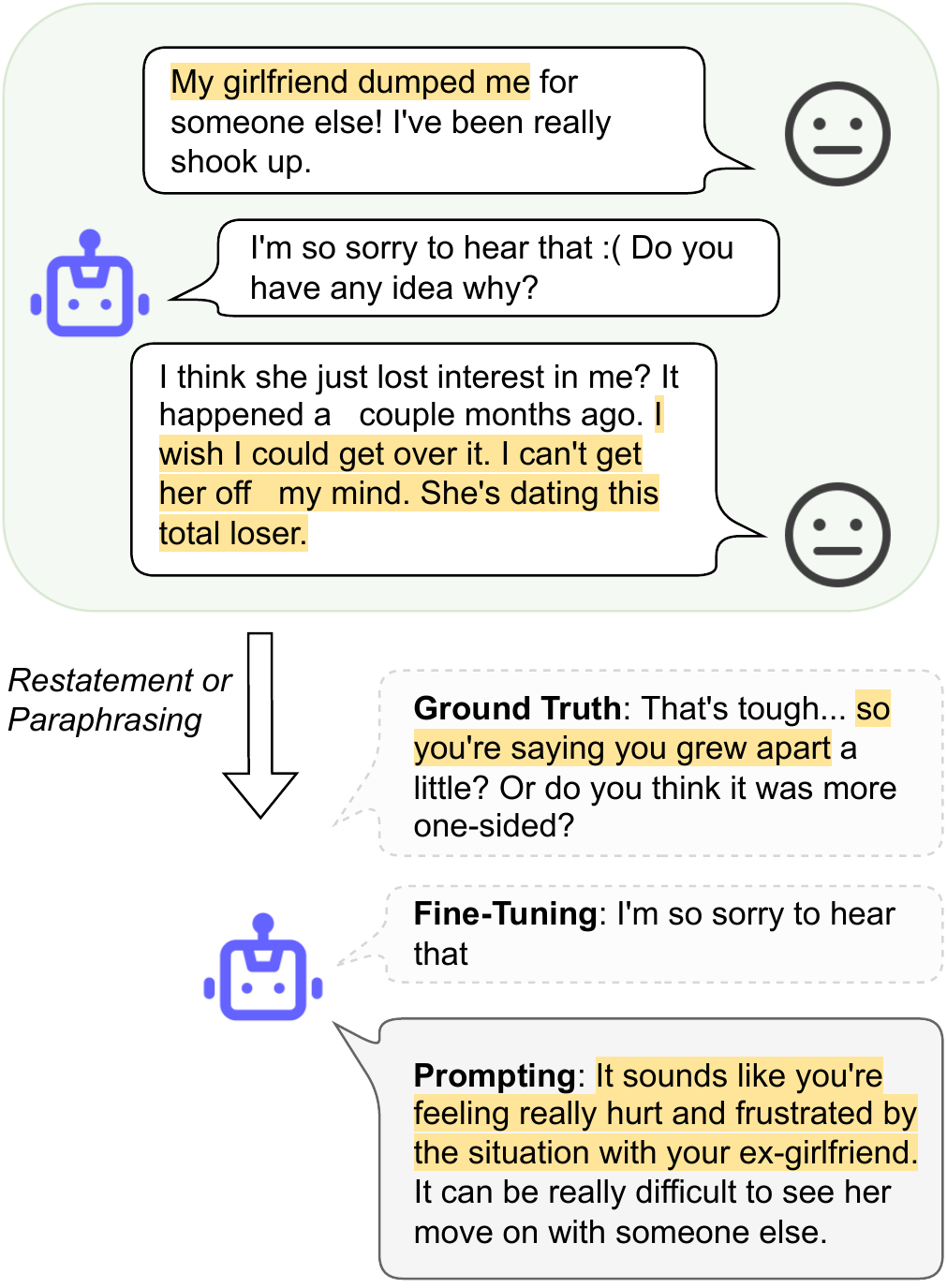}}
     \caption{Excerpt of a conversation between an emotional help-seeker and a supporter about a breakup, with candidate responses attempting to use the support strategy ``Restatement or Paraphrasing.'' 
     }
     \label{fig:ComparisonExample}
 \end{figure}
 
 Mixed initiative dialogue systems allow all interacting agents to initiate actions to control the interaction. These systems dynamically adapt interaction styles to regain control and progress towards specific goals~\cite{allen1999mixed,chu2000mimic}, unlike others which passively respond to users' input (e.g. some assistants like ChatGPT),  
 
 Mixed initiative dialogue systems thus often involve complex policy planning sub-tasks to determine optimal turn-level system dialogue intents~\cite{peng2018deep,hiraoka2013dialogue,muise2019planning,liu2020towards}. These policies define when it is optimal for a system to regain initiative (e.g., when a moderator should interject in a conversation, or when a companion should ask questions or change a conversation topic). 
 
 However, ``optimal'' planned dialogue intents still need to be executed through ``optimal'' response models. The standard practice in recent dialogue research has been to fine-tune a pre-trained language model for conditional generation to achieve semantic control through some combination of innovations in model architectures or learning processes~\cite{liu2021towards,chen2019semantically}. Such generation approaches still leave room for error. Assuming that there exists a truly optimal dialogue policy planner, a response model may still generate according to the wrong intent (partially due to the fact that dialogue datasets often have annotation errors~\cite{qian2021annotation,zang2020multiwoz}). Or, a model may learn to generate correct intents but fail to create a response consistent with conversational context~\cite{chen-etal-2022-seamlessly}. Additionally, training corpora often differ in demographic and distribution compared to production environments, which can lead to deteriorating response quality~\cite{koh2021wilds}.
 
 We propose using vanilla large pre-trained language models (LLMs) such as GPT-3~\cite{brown2020language} as drop-in replacements to traditional fine-tuned conditional generation models for mixed-initiative dialogue systems. LLMs typically have been trained on massive corpora with large amounts of linguistic variety, making them more robust to overfitting specific tasks. Recent work demonstrates that LLMs have reasonable semantic control through few-shot prompting~\cite{brown2020language,chen2023places,meng2022generating}. Here, we demonstrate how\footnote{Code to reconstruct all prompts forthcoming. It will be released at https://github.com/maxlchen/Controllable-Mixed-Initiative-Dialogue-Generation} to systematically prompt LLMs for mixed-initiative dialogue generation. Evaluations yielded strong performance on two popular English mixed-initiative tasks: Emotional Support Conversations (ESC; \citet{liu2021towards}) and PersuasionForGood (P4G; \citet{wang-etal-2019-persuasion}).
 \section{Related Work}
 \paragraph{Controllable Generation}
  approaches often involve fine-tuning a model conditioned on control codes \cite{ctrl, ficler-goldberg-2017-controlling}, additional attribute representations in hidden states \cite{hoang2016incorporating, fu2018style} or latent variables \cite{bowman2016generating, wang-etal-2019-topic}. Other work has attempted to mitigate the computational cost of fine-tuning, e.g. by training an auxiliary networks to guide the original LM \cite{Dathathri2020Plug, yu-etal-2021-attribute-alignment, pascual-etal-2021-plug-play}. Here, we attempt controllable generation that replaces fine-tuning by prompting LLMs.
 
 \paragraph{Prompting in Dialogue Research} typically has focused on understanding tasks such as dialogue planning~\cite{kuo2022zero} or state tracking~\cite{lee2021dialogue,mi2022cins}. More recent dialogue research has examined using prompting for generating conversational data with varying levels of control~\cite{kim2022soda,chen2022weakly,mehri2022lad,chen2023places}, citing the difficulty of using vanilla language models in production. Studies focusing on response generation looked at prompting LLMs specifically for knowledge-grounded dialogue generation \cite{liu2022multi,madotto2021few,shuster2022language}. Our work is the first to construct an interactive prompt-based mixed initiative dialogue system and evaluate the semantic control of prompting.
 \section{Datasets}
 We examined ESC \cite{liu2021towards}) and P4G \cite{wang-etal-2019-persuasion}. ESC consists of 1053 conversations between emotional help-seekers and supporters. Each conversation is annotated with the help-seeker's description of their problem, and the type of issues they are facing. Each turn by the supporters is annotated with one of eight emotional support strategies (Table~\ref{ESConv_Prompts}). P4G contains 300 annotated conversations between persuaders who attempt to persuade persuadees to donate to a charity called Save the Children. Persuader turns are annotated with one of 10 strategies (Table~\ref{P4G_Prompts}). 
 \section{Baselines}
 In mixed-initiative dialogue, interacting parties continuously exchange control throughout the conversation. However, in order for agents to regain control, they must be able to properly execute items from their conversational agenda, e.g. generating a response that matches a desired strategy/intent. 
 
 \citet{liu2021towards} fine-tuned BlenderBot~\cite{roller2021recipes} on ESC using input representations consisting of flattened dialogue history and the predicted emotional support strategy for a specific turn. The best-performing model in their experimental setting is ``Oracle-BlenderBot'' which conditions on the ground truth strategy for a given turn.
 
 \citet{chen-etal-2022-seamlessly} proposed a persuasive dialogue system called RAP, which combined targeted user response with conditional generation. The conditional generation component of RAP involves fine-tuning BART~\cite{lewis2020bart} using a penalized loss to force the model to artificially create semantic control through dialogue intents. 
 \section{Mixed-Initative Dialogue Prompting}
 RAP required introducing a dialogue intent classifier to weakly supervise the training process, as there is not an oracle for whether the dialogue intent of a candidate response is correct. But, this confounds errors, as classifiers are imperfect. Moreover, fine-tuning approaches like both RAP and Oracle-BlenderBot involve balancing a tradeoff between response quality and semantic control accuracy. Prompting LLMs avoids both issues as it does not involve adjusting model weights to learn representations of control codes for individual tasks. 
 
 In this paper, we systematically prompt InstructGPT ``text-davinci-003.'' Rather than requiring expert-level prompt engineering, we create general prompt templates which directly fill slots using roles and annotations from both ESC and P4G. Specifically, we split up prompt construction into \textit{Task Background} and \textit{Conversation History}.
 
\begin{figure}[t]
     \centering
     \scalebox{0.90}{\includegraphics[width=\linewidth]{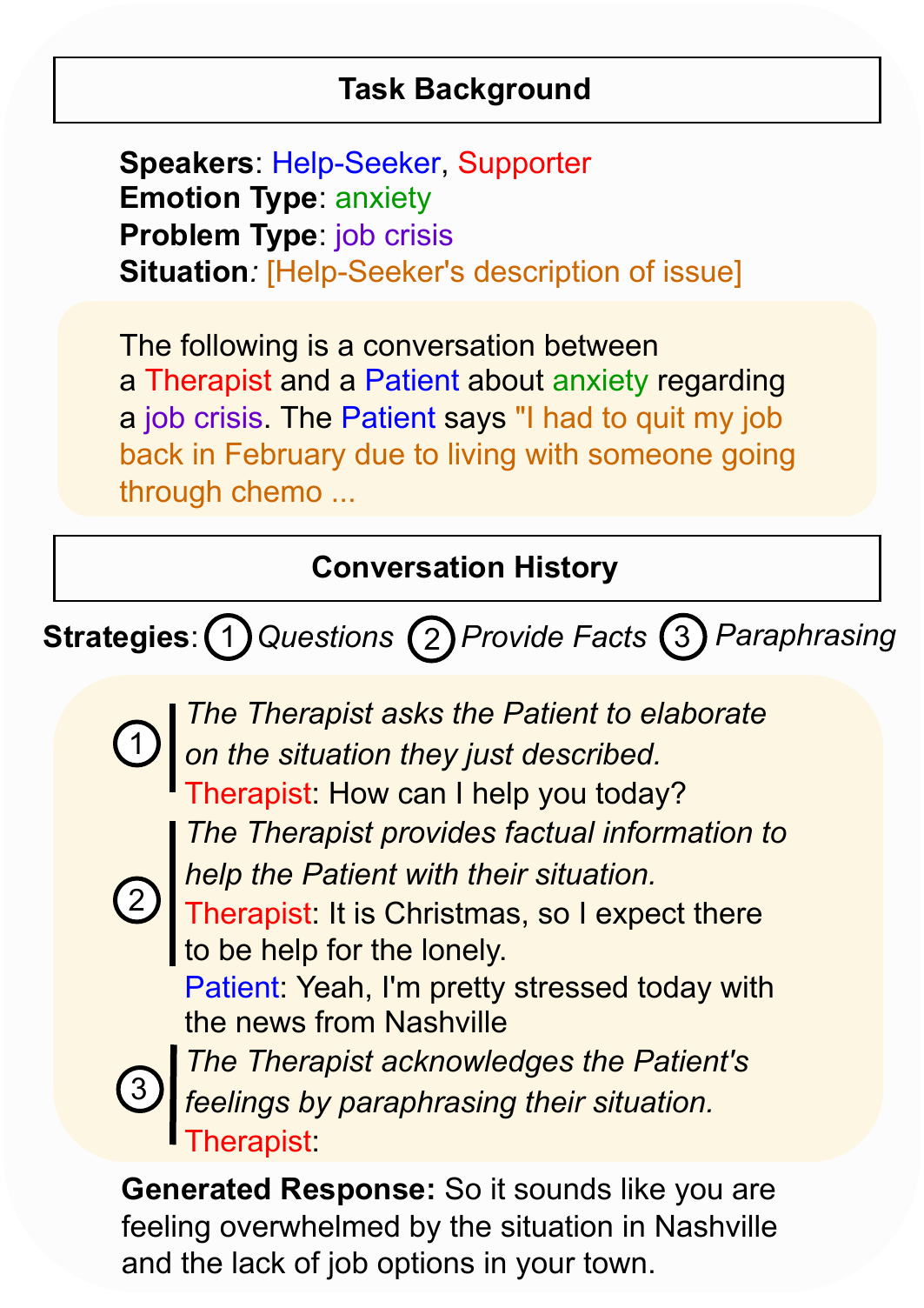}}
     \caption{Parts of an example prompt for ESC (yellow background). Task Background: ground truth annotations describing the conversation. Conversation History: dialogue context with natural language forms of annotated dialogue intents. Full situation in Appendix ~\ref{additionalPromptDetails}.}
     \label{fig:ESConvPrompt}
 \end{figure}

 Figure~\ref{fig:ESConvPrompt} breaks down an example of a prompt for ESC. The Task Background is a paragraph formed from the ``emotion type,'' ``problem type,'' and ``situation'' annotations provided by the corpus. The Conversation History consists of each prior utterance, prepended by labels for each speaker. The system-side turns are also prefixed by a natural language form of the annotated emotional support strategy, derived from the annotation scheme in \citet{liu2021towards} (e.g. ``The Therapist acknowledges the Patient's feelings by paraphrasing their situation.''). Figure~\ref{fig:ESConvPrompt} contains the contextual dialogue turns in order, along with the three support strategies used.
 
 The P4G prompting style is similar. Unlike personalized emotional support conversations, the task does not change, so the Task Background is fixed with relevant factual background information. The Conversation History still interweaves narrative directions for each persuasive strategy (e.g. ``The Persuader uses a logical appeal.''). Example provided in Figure~\ref{fig:P4GPrompt}. The natural language intent mappings for both tasks are provided in Tables~\ref{ESConv_Prompts},\ref{P4G_Prompts}.
 \begin{table}[t]
 \small
 \begin{center}    
 \begin{tabular}{@{}lllll@{}} \toprule
 Corpus & Metric & FT & GT & Prompt \\ \midrule
 \multirow{10}{*}{\textbf{ESC}} & Accuracy & 0.81 & 0.85 & \textbf{ 0.88$^{*}$}  \\
 & Coherence &  3.57 &  3.57  &   \textbf{ 3.72}      \\
 & Consistency &  3.63  &  3.60 &  \textbf{ 3.80$^{+*}$}        \\
 & Engagingness &  3.55 &  3.61 &   \textbf{ 3.81$^{+*}$}   \\
 & Distinct-3 & 0.89 & 0.90 & \textbf{ 0.90} \\
 & Distinct-4 & 0.87 & 0.90$^{*}$ &\textbf{ 0.91$^{+*}$} \\
 & QuantiDCE & \textbf{3.25} & 3.03 & ~3.19 \\
 & \textit{Win Rates} \\
 & ~~~~~~v. FT & & \textbf{0.56} & 0.52 \\
 & ~~~~~~v. GT & 0.44 & & \textbf{0.64$^{*}$} \\
 & ~~~~~~v. Prompt & 0.48 & 0.36 & \\
 \midrule
 \multirow{10}{*}{\textbf{P4G}} & Accuracy & 0.88  & 0.83 & \textbf{ 0.89} \\
 & Coherence & 3.66    & 3.58 &  \textbf{ 3.83$^{+*}$}       \\
 & Consistency & 3.69   &  3.56 &  \textbf { 3.71$^{+}$}  \\
 & Engagingness & 3.62   & 3.52 &  \textbf{ 3.69$^{+}$}     \\
 & Distinct-3 & 0.87 & 0.88 & \textbf{ 0.89} \\
 & Distinct-4 & 0.88 & 0.88 & \textbf{ 0.88} \\
 & QuantiDCE & 3.16 & 3.09 & \textbf{ 3.24$^{+}$} \\
 & \textit{Win Rates} \\
 & ~~~~~~v. FT & & 0.56 & \textbf{0.59$^{*}$} \\
 & ~~~~~~v. GT & 0.48 & & \textbf{0.55} \\
 & ~~~~~~v. Prompt & 0.41 & 0.45 & \\
 \bottomrule
 \end{tabular}
 \end{center}
 \caption{Evaluation of response quality and semantic control accuracy. 
 FT: fine-tuning (Oracle-BlenderBot for ESC; RAP for P4G). GT: ground truth utterances. At $\alpha=0.05$: $^+$ is greater than ground truth and $^*$ is greater than fine-tuning.}
 \label{StaticEvaluation}
 \end{table}
 \section{Experiments}
 We evaluated prompting statically and interactively.
 \subsection{Static Evaluation}
 We quantified how much semantic and pragmatic control vanilla LLMs can provide in conversation. We randomly sampled 100 responses from ESC (supporters) and P4G (persuaders). Each response's conversational history and strategy annotation was used to generate responses via prompting and fine-tuned models. We used Oracle-BlenderBot for ESC and RAP's conditional generation module for P4G.
 
 We asked crowdworkers on Amazon Mechanical Turk\footnote{Details for all human evaluation tasks in Appendix~\ref{HumanEvalDetails}.} to evaluate candidate responses' accuracy with respect to its prescribed dialogue intents, coherence, consistency, and engagingness. We paired the dialogue responses from each source (fine-tuning, prompting, or ground truth) with the corresponding responses from each of the other sources, allowing us to compute preference win-rates between each pair. Each job presented only one pair of responses, in a random order. Additionally, we examined automatic metrics through Distinct-N ($N \in \{3,4\}$), as well QuantiDCE~\cite{ye2021towards}, a BERT-based automatic dialogue coherence metric for open-domain conversation.
 
 Table~\ref{StaticEvaluation} shows that prompt-generated responses are more highly rated in terms of quality compared to responses generated from competitive fine-tuned dialogue models \textit{as well as ground truth responses}, in terms of all human evaluation metrics. This is also the case for Distinct-N in both tasks, and QuantiDCE in P4G. Oracle-BlenderBot slightly outperforms the prompt-generated responses in terms of QuantiDCE for ESC, but this difference is not statistically significant. Table~\ref{StaticEvaluation} also shows that the prompt-generated responses are consistently preferable to the responses generated from fine-tuned dialogue models as well as the ground truth. 
 
 Finally, we also see that prompting appears to provide the best semantic control over generated responses. Prompt-generated responses had the highest probability of matching the desired dialogue intent, even surpassing that of the ground truth utterances in both corpora. This further demonstrates the difficulty of performing annotation for supervised training --- the conversational strategies are subjective, and even the ground truth responses may have annotation errors. The prompt-generated responses are generally of higher quality than both fine-tuned models, which may be a result of the aforementioned difficulty of balancing control accuracy with response quality during generation.
 \subsection{Interactive Evaluation}
 We evaluated prompting as a generation module for mixed-initiative systems. This requires holding fixed other components, including policy planning. RAP is a recently proposed framework for P4G using an ``optimal'' persuasive strategy ordering. But, it built rapport with users by hierarchically integrating social chit-chat and knowledge retrieval with semantically-controlled generation (details in \citet{chen-etal-2022-seamlessly}). We built a system which replaces RAP's fine-tuned BART module with a module that systematically prompts InstructGPT. 
 As with the original implementation of RAP, our prompting module conditions on the knowledge retrieved for factual question answering\footnote{Implementation details in Appendix~\ref{implementationDetails}.}. 
 
 We asked crowdworkers to evaluate our system according to the criteria in Table~\ref{InteractiveEvaluation}. The system using prompting for generation was consistently rated more favorably than RAP, including in terms of convincingness, persuasiveness, and being a strong reason for donation. We discuss conversation examples in Appendix~\ref{sec:conversation_examples}. We see that our system was robust to a variety of input language patterns.
 \begin{table}[t]
 \centering
 \scalebox{0.80}{
 \begin{tabular}{llll}
 \toprule
 \textbf{The chatbot...}                                        & \textbf{RAP (FT)} & \textbf{Prompting} \\ \midrule
 is competent      $\uparrow$ & 3.81$\pm$1.11   & \textbf{4.21$\pm$0.84$^{**}$}   \\ 
 is natural        $\uparrow$ & 3.81$\pm$1.19  & \textbf{4.17$\pm$0.94}  \\ 
 is intelligent     $\uparrow$ & 3.83$\pm$1.20   & \textbf{4.19$\pm$1.05}   \\ 
 is well-intentioned   $\uparrow$ & 4.00$\pm$1.09   & \textbf{4.29$\pm$0.87}   \\ 
 is confident      $\uparrow$ & 3.94$\pm$1.13 & \textbf{4.35$\pm$0.85$**$}   \\ 
 was dishonest   $\downarrow$ & 2.90$\pm$1.42   & \textbf{2.70$\pm$1.40} \\ 
 is warm  $\uparrow$ & 3.56$\pm$1.31 & \textbf{4.04$\pm$1.00$^{**}$}  \\ 
 is sincere  $\uparrow$ & 3.85$\pm$1.25 & \textbf{4.25$\pm$0.90$^{*}$}  \\ 
 is efficient  $\uparrow$ & 3.96$\pm$1.18 & \textbf{4.33$\pm$0.75$^{*}$}  \\ 
 tried to pressure me  $\downarrow$ & 3.04$\pm$1.39   & \textbf{3.02$\pm$1.23} \\
 increased my intent to donate  $\uparrow$ & 4.00$\pm$1.07   & \textbf{4.15$\pm$0.84} \\ 
 is persuasive      $\uparrow$ & 3.83$\pm$1.14  & \textbf{4.06$\pm$1.06}  \\ 
 is convincing     $\uparrow$ & 3.77$\pm$1.14   & \textbf{4.29$\pm$0.73$^{**}$}    \\ 
 is a strong reason for donating  $\uparrow$ & 3.60$\pm$1.30 & \textbf{4.19$\pm$0.81$^{**}$}  \\ 
 \bottomrule
 \end{tabular}
 }
 \caption{Comparison of chatbots using RAP with fine-tuning and prompting on the interactive P4G task. Results are $\mu \pm \sigma$, scale is 1 to 5. $^{**}$ indicates significance at $\alpha=0.05$, $^{*}$ indicates significance at $\alpha=0.10$. 
 }
 \label{InteractiveEvaluation}
 \end{table}
 \section{Discussion}
 \label{sec:discussion}
 Prompting yields strong performance in mixed-initiative tasks in the low resource regime\footnote{We prompt without full conversation examples in-context.}. Prompt-generated responses are often preferable even compared to ground-truth responses in ESC and P4G. From 17 paired evaluations of ESC where crowdworkers rated ground truth utterances as not matching the ground truth intent annotation, the prompt-generated response was rated as correct 13 times. However, this is likely because many dialogue corpora are created or annotated by crowdworkers, so the data may vary in quality. While LLMs may generate ``better'' responses than crowdworkers, we \textit{do not} expect them to be better than expert therapists.
 
 The results \textit{do} indicate that prompting may be appropriate for building systems for tasks with limited data. As made evident by our ratings, annotating dialogue intents is a difficult and subjective process prone to errors \textit{which can further propagate to fine-tuned task models}. This could potentially be addressed by the high semantic control demonstrated through prompting, despite not requiring downstream fine-tuning label supervision. 
 
 This prompting approach could be applied to other mixed-initiative tasks, including chit-chat and task-oriented dialogue. 
  For instance, many real-world systems such as customer service chatbots already have pre-defined policies for what systems are allowed to say, despite not necessarily having many labeled conversations. A system can be designed as long as there is a policy planner, which could simply be a hierarchical ruleset. While there is some human-effort involved in writing natural language forms of fixed dialogue intents, it is a much less costly process than annotating high-quality dialogue data.
 \section{Conclusion}
 We find encouraging results for prompting on mixed-initiative dialogue tasks, indicating that generated responses are high quality and follow semantic controls. Strong low resource performance opens the possibility of future work building mixed-initiative systems around novel settings which would require subjective data annotation. 
 
 \section{Limitations}
 \paragraph{Limits of Prompt-based Generation.}
 This work specifically proposes improvements to the controllable generation portion of mixed-initiative dialogue systems. However, dialogue policy planning is still an important problem to consider. In order to evaluate generation improvements, we hold dialogue policies fixed --- in the static evaluation, we condition on ground truth dialogue intents, and in the interactive evaluation, we follow the same dialogue intents prescribed by the RAP system. To this end, a mixed-initiative dialogue system \textit{cannot consist solely} of a generation module powered by prompting. There needs to be a set of rules or models that govern how a system can regain control of a conversation; the generation module is just a means of enacting these rules. As discussed in Section~\ref{sec:discussion}, prompting is a great option if there is already a pre-existing policy planner. 
 
 Due to these limitations, we did not conduct an interactive evaluation in the ESC setting. Emotional support conversations are highly personal, as circumstances vary across individuals. It would have required having study participants pretend to require support regarding a fixed scenario, or for participants to disclose their personal issues, which can raise other ethical concerns. Moreover, dialogue policy planning is not straightforward for emotional support, due to this highly variable nature. Effective support strategy planning requires expert knowledge. 
 
  In Section~\ref{sec:discussion}, we also discussed that prompting may be appropriate for developing systems for novel tasks in low-resource settings. However, deploying prompt-based systems may be less useful for the purpose of setting new benchmarks on existing leaderboards with a plethora of data. Such setting already have plenty of well-annotated conversations and simple fine-tuned models can often achieve strong performance.
 \paragraph{Guardrails.}
 Proper guardrails should be put in-place prior to productionization of any dialogue system, prompt-driven or not. While we witness strong overall response quality both in terms of human evaluation and automatic metrics, language models can generate contradictions. System builders may consider employing guardrails for dialogue consistency (e.g. \citet{jin2022improving}) and coherence (e.g. \citet{ye2021towards}), among others. 
 
 As with any training set, InstructGPT and other  LLMs have been trained on finite amounts of data. InstructGPT has not been trained on data after 2021. This is also true of training corpora such as P4G or ESC; these corpora were published in 2019 and 2021, respectively. Particularly in any sensitive environments, guardrails should be put in-place for factual correctness (e.g. \citet{santhanam2021rome,wang2020improving}). RAP attempted to remedy this by incorporating retrieval for factual questions, which we also embedded into our prompting approach, but this knowledge base is also finite. In Section~\ref{sec:conversation_examples} we discuss one such example (Table~\ref{negative_example_1}). A possible solution is internet retrieval~\cite{komeili2022internet}, but search engines can also yield misinformation, which leads to hallucination.
 
 \paragraph{Computational Cost of Language Models.}
 LLMs are computationally expensive, and in the case of models such as InstructGPT, they are not open source. However, in this study, we did not have access to equally powerful open-source models such as OPT 175B, nor the appropriate hardware to load such a model (loading OPT 175B requires 350 GB of GPU memory). We performed initial experiments with much smaller models which fit our hardware constraints such as GPT-J 6B, but there was much higher variance in performance. This is supported by the fact that many reasoning capabilities do not seem possible with models smaller than 175B parameters~\cite{wei2022chain,wei2022emergent}. Given our limited budget for human evaluation, we opted to use the best performing LLM we had access to, InstructGPT.
 
 \paragraph{Prompt Optimality}
 It is possible that we do not use an ``optimal'' set of prompts as we did not mine prompts or perform soft prompting. However, prompt optimality itself is a problem in dialogue generation, because open-ended dialogue evaluation is a difficult task. Most automatic evaluation metrics do not align well with human ratings in dialogue~\cite{yeh2021comprehensive,liu2016not}. This makes it suboptimal to use as a discriminator in soft prompting, for instance. Most existing work that does search for optimal prompts or tunes prompts works with tasks that have clearly defined automatic evaluation, such as sentiment analysis or table-to-text generation~\cite{van2022don,li2021prefix,lester2021power}. Moreover, human ratings are expensive and not scalable for systematic optimization.
 
 \section{Ethics Statement}
 \paragraph{Chatbot Identities.}
 All study participants were informed that they were speaking to a chatbot, in accordance with law in certain localities (e.g. California's Bot Disclosure Law). 
 \paragraph{Dangers of Fully Automated Dialogue Systems.}
 We do not encourage the deployment of fully automatic dialogue systems for tasks such as emotional support in production settings. Bot Disclosure Laws exist because knowledge of chatbot identities affect human perception \cite{shi2020effects}, and thus in sensitive situations such as therapy or emotional support, patients may not receive adequate support. Moreover, there is the possibility of emotional support dialogue systems without proper guardrails introducing harmful or otherwise unethical content, e.g. by mentioning references which could be considered ``triggering.'' Instead, we advise the use of mixed-initiative dialogue systems in a supportive manner, e.g., to assist trained counselors who have the emotional intelligence to recognize what content may be hurtful.
 \paragraph{Reproducibility.}
 In this study we used GPT-3, which is not an open-access language model. However, we have clearly described all of the prompts used in our paper.
 
 \paragraph{Data Biases} Every dataset, including P4G and ESC, has its own biases. LLMs such as InstructGPT have been trained on large amounts of data but may still not capture language usage of a sufficiently diverse population. While in Appendix~\ref{sec:conversation_examples} we see InstructGPT's ability to handle diversity in language, this is something that warrants further interactive study with more extreme cases.
 
 \paragraph{Crowdsourcing.} 
 All crowdworkers were paid at a rate of \$15 per hour. We did not collect any personal or demographic information about any workers. Our study and  data collection process has received IRB approval.
 \section*{Acknowledgements}
 Thanks to Intel for supporting this work through a research gift. We thank Ta-Chung Chi, Kun Qian, and our anonymous peer-reviewers for their helpful feedback. We also thank Sophie Chen for helpful suggestions on designing our figures. 
 \bibliography{anthology,custom}
 \clearpage
 \appendix
 \setcounter{table}{0}
 \renewcommand{\thetable}{A\arabic{table}}
 \setcounter{figure}{0}
 \renewcommand{\thefigure}{A\arabic{figure}}
 \begin{figure}[ht]
     \centering
     \scalebox{0.95}{\includegraphics[width=\linewidth]{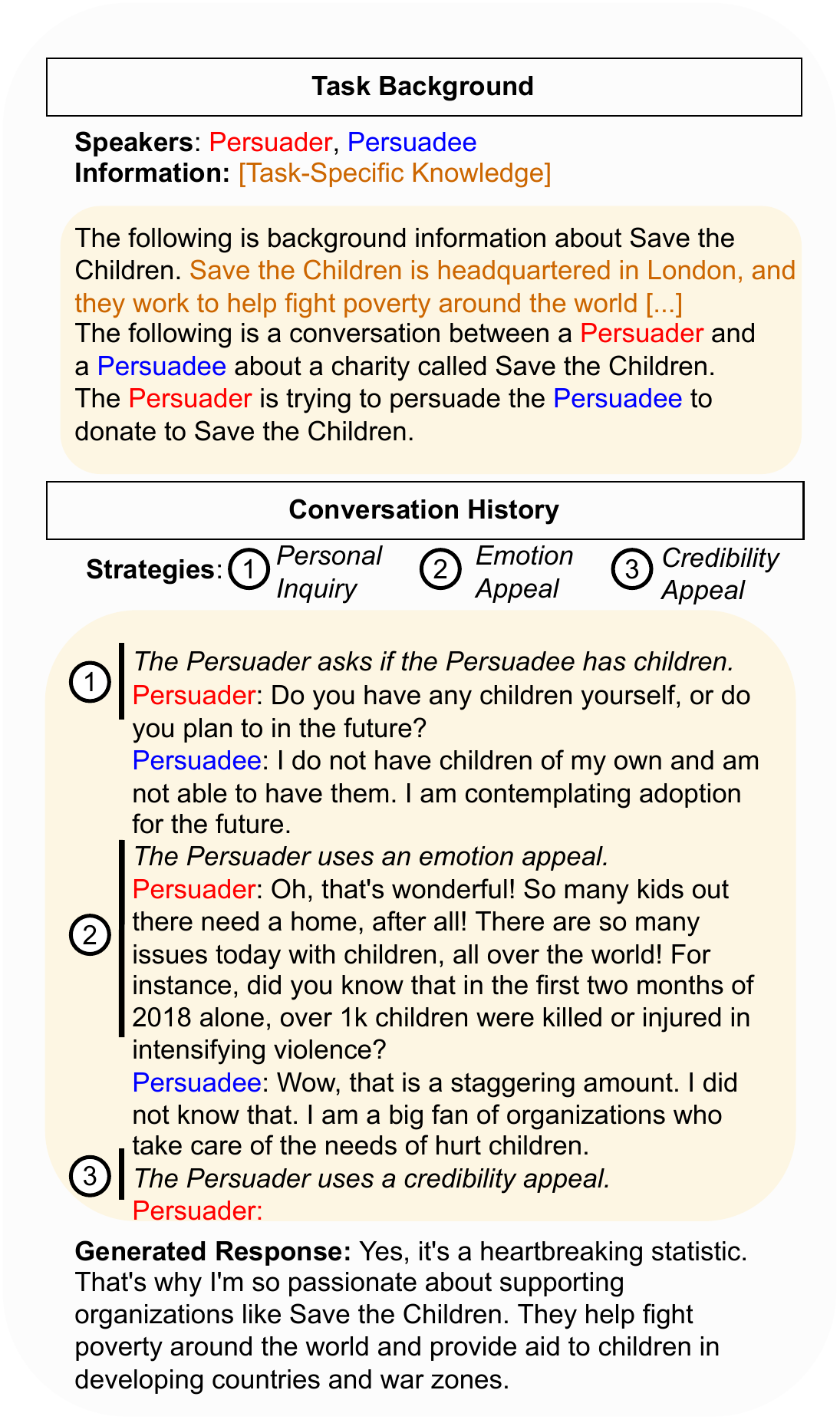}}
     \caption{Example prompt broken into two parts for P4G. Task Background is knowledge about Save the Children. The Conversation History consists of dialogue turns interwoven natural language forms of the Persuader's annotated dialogue intents. The full Task Background (including task-specific knowledge) used for P4G is given in Appendix \ref{additionalPromptDetails}.}
     \label{fig:P4GPrompt}
 \end{figure}
 \section{Human Evaluation Details}
 \label{HumanEvalDetails}
 We performed both our static and interactive evaluation on Amazon Mechanical Turk. We required that all crowdworkers had a HIT Approval Rate of at least $95\%$. 322 unique crowdworkers successfully completed the static evaluation task. There were 100 unique conversation turns used, with each candidate response being rated twice in order to pair the three conditions (ground truth, fine-tuning, prompting). 100 unique crowdworkers successfully completed the interactive evaluation task.
 
 For the static evaluations of both ESC and P4G, the following definitions were provided to the crowdworkers:
 
 \begin{itemize}
     \item Engaging (1-5): Whether the response is interesting and engaging.
     \item Coherent (1-5): Whether the response makes sense and is non-repetitive.
     \item Consistent (1-5): Whether the response is free of inconsistencies and logical fallacies.
 \end{itemize}
 
 Specifically for P4G, the following conversational strategies were defined along with examples:
 \begin{itemize}
     \item Greeting: A greeting from the speaker.
     \item Source-related inquiry: A question about the charity, Save the Children.
     \item Task-related inquiry: A question related to the task of donating to Save the Children, e.g. asking whether the Persuadee has donated to charities in the past or asking about information related to Save the Children.
     \item Personal-related inquiry: A personal question about the persuadee.
     \item Credibility appeal: An argument giving credibility to Save the Children.
     \item Emotional appeal: An argument that elicits an emotional response from the Persuadee.
     \item Logical appeal: An argument that uses reasoning and evidence to convince the Persuadee, e.g., by using facts to reason that a donation would make a tangible impact.
     \item Self-modeling: A reflection of the Persuader's own intention to donate to Save the Children.
     \item Foot-in-the-door: A strategy of starting with small donation requests to facilitate compliance followed by larger requests.
     \item Personal story: Using narrative examples relating to the Persuader's personal experiences or other anecdotes.
     \item Propose donation: Asking the Persuadee if they would like to donate to the charity.
     \item Closing: Ending the conversation.
 \end{itemize}
 For ESC, the following support strategies were defined along with examples:
 \begin{itemize}
     \item Question: The Therapist asks the Patient for information to help them articulate their issues.
     \item Restatement or Paraphrasing: A simple, concise rephrasing of the help-seeker's statements.
     \item Reflection of Feelings: Acknowledge/articulate and decsribe the help-seeker's feelings.
     \item Self-disclosure: The Therapist divulges similar experiences they have had.
     \item Affirmation and Reassurance: Affirm the Patient's strengths, motivation, and capabilities and provide reassurance and encouragement.
     \item Providing suggestions: Provide suggestions about how to change.
     \item Information: Provide useful information, often backed with data, facts, or opinions.
     \item Others: Exchange pleasantries and use other support strategies not listed above.    
 \end{itemize}
 The persuasion strategies are defined based on \citet{wang-etal-2019-persuasion}, and the emotional support strategies are defined based on \citet{liu2021towards}.
 \begin{table*}[t]
 \small
 \begin{tabular}{p{0.25\linewidth}|p{0.74\linewidth}}
 Dialogue Intent & Natural Language Form \\ \hline
 Question &  The Therapist asks the Patient to elaborate on the situation they just described. \\
 Self-disclosure & The Therapist provides a statement relating to the Patient about the situation they just described. \\
 Affirmation and Reassurance & The Therapist provides affirmation and reassurance to the Patient on the situation they just described. \\
 Providing Suggestions & The Therapist provides suggestions to the Patient on the situation they just described. \\
 Others &  \\
 Reflection of feelings & The Therapist acknowledges the Patient's feelings about the situation they described. \\
 Information & The Therapist provides factual information to help the Patient with their situation. \\
 Restatement or Paraphrasing & The Therapist acknowledges the Patient's feelings by paraphrasing their situation. \\
 \end{tabular}
 \caption{Mapping of Supporter conversational strategies to natural language in Emotional Support Conversations.}
 \label{ESConv_Prompts}
 \end{table*}
 \begin{table*}[t]
 \small
 \begin{tabular}{p{0.25\linewidth}|p{0.74\linewidth}}
 Dialogue Intent & Natural Language Form \\ \hline
 Personal Story & The Persuader tells a personal story.  \\
 Credibility Appeal & The Persuader uses a credibility appeal.  \\
 Emotion Appeal & The Persuader uses an emotion appeal.  \\
 Propose Donation & The Persuader asks if the Persuadee would like to make a small donation.  \\
 Foot-in-the-door &  The Persuader tells the Persuadee about how useful even small donations are.  \\
 Logical Appeal &  The Persuader uses a logical appeal.  \\
 Self-modeling &  The Persuader talks about how often they donate to charities.  \\
 Task-related inquiry &  The Persuader asks the Persuadee if they have donated to any charities before.  \\
 Source-related inquiry & The Persuader asks the Persuadee if they have heard of Save the Children before.  \\
 Personal-related-inquiry & The Persuader asks the Persuadee if they have kids.
 \end{tabular}
 \caption{Mapping of Persuader dialogue intents and conversational strategies to natural language in Persuasion for Good.}
 \label{P4G_Prompts}
 \end{table*}
 For the interactive evaluation, all crowdworkers were randomly assigned a link to a chatbot running either RAP or a prompt-driven system deployed using the LegoEval platform~\cite{li2021legoeval}. In total, 48 crowdworkers used the prompt-based system, and 52 crowdworkers used the system powered by RAP after removing those who did not successfully answer the validation question. All crowdworkers agree to interacting with a research prototype which may produce harmful content. They also were required to provide content to the logging of their responses and ratings.
 \section{Implementation Details}
 \label{implementationDetails}
 All baseline models were trained using HuggingFace Transformers~\cite{wolf2020transformers} and PyTorch~\cite{paszke2019pytorch}. All experiments used one NVIDIA A6000 GPU.
 
 The rest of the RAP baseline follows the details provided in \citet{chen-etal-2022-seamlessly}. To perform knowledge retrieval,  we computed the cosine distance of Sentence-BERT \cite{reimers-gurevych-2019-sentence} embeddings between question-answer mappings derived from the training data, and retrieved the answer to the question that has the lowest cosine distance in semantic meaning from the question asked by the user. In order to use the knowledge in our prompts, we simply append the retrieved knowledge to the end of the prompt. For example, the prompt typically ends with an indicator that the Persuader should speak --- "Persuader:". Now, the prompt instead ends with "Persuader: [retrieved knowledge]".
 
 In RAP, the authors used Blender Bot 2.0~\cite{xu2022beyond,komeili2022internet} to incorporate social chitchat in order to acknowledge user responses. In our version using prompting for generation, we directly add more instructions into the prompt. We prepend the natural language form of the system-side dialogue intent with ``The Persuader acknowledges the Persuadee's response and''. For example, a prompt targeting generating a credibility appeal with social acknowledgement would be ``The Persuader acknowledges the Persuadee's response and The Persuader uses a credibility appeal.''
 
 \subsection{Additional Prompt Details}
 \label{additionalPromptDetails}
 The full situation given in the prompt example from Figure~\ref{fig:ESConvPrompt} is as follows:
 \textit{``I had to quit my job back in February due to living with someone going through chemo. My town doesn't have many job options other than retail, so I have been trying to earn money for debts online.''}
 
 The full Task Background for P4G is as follows:
 \textit{``The following is background information about Save the Children. Save the Children is headquartered in London, and they work to help fight poverty around the world. Children need help in developing countries and war zones. Small donations like \$1 or \$2 go a long way to help. \\ The following is a conversation between a Persuader and a Persuadee about a charity called Save the Children. The Persuader is trying to persuade the Persuadee to donate to Save the Children.''}
 
 Prompting InstructGPT for P4G cost $\$0.06$ per study participant, on average. We generate using a temperature of 0.70, and frequency penalty of 0.75. Our prompting code is attached and will be made available online upon acceptance.
 \section{Example Conversations \& Case Study}
 \label{sec:conversation_examples}
 
 Table~\ref{positive_example_1} and Table \ref{positive_example_2} are examples of users who agreed that the prompt-based chatbot was both persuasive and increased their intention to donate. They also both found that the chatbot created natural and coherent responses.  The user in Table \ref{positive_example_2} thought that the chatbot's responses were also very logically consistent, but the user in Table \ref{positive_example_1} provided a neutral opinion. 
 
 In Table \ref{positive_example_1}, the user appears engaged from the start. However, they reveal an interest in whether Save the Children is active in Brazil, and admit that they are from Brazil. InstructGPT is able to generate responses which correctly identify that Save the Children is indeed active in Brazil, and able to form coherent anecdotes about this topic. Similarly, the user in Table \ref{positive_example_2} appears to warm up to the chatbot throughout the conversation. By their fifth turn, they actually admit ``i think i would be interested in making a donation'' and their responses are more verbose as the conversation continues. 
 
 On the other hand, the users in Table \ref{negative_example_1} and Table \ref{negative_example_2} both disagreed with the statement that ``The chatbot is persuasive.'' However, the actual conversation context leading to these statements is quite different. In Table \ref{negative_example_1}, the user seems actively engaged throughout the conversation. They ask several questions, and each time, the system generates a reasonable response. For instance, on Turn 4, the user asks ``Are they involved in the situation in Ukraine?'' and the system correctly responds, ``Yes, they are. Save the Children has been present in Ukraine since 2014 ...'' The user even rates the chatbot responses as natural, coherent, and consistent, but they did not find it to be persuasive. This perhaps speaks to the importance of having personalized persuasive strategy planning, which is an open problem in P4G. Additionally, it should be noted that while the user is likely making reference to the ongoing war in Ukraine at the time of writing, InstructGPT is likely more generally referring to the fact that Save the Children has been present in Ukraine since 2014\footnote{https://www.savethechildren.net/what-we-do/emergencies/ukraine-crisis} which would be a fact that exists in its publicly available pre-training corpora dating back to 2021. While it was not an issue in this scenario, future work may look to consider expanding the Task Background in the prompt to further include critical information to condition on. Additionally, as LLMs continue to be empowered with internet retrieval, up-to-date information retrieval will be more feasible in such applications.
 
 Table \ref{negative_example_2} is quite different from the other examples displayed. The user looks disengaged from the task, as their responses are provided in all capitals. They do not provide more than two words in a single response, and their responses often are incoherent or contain typos. They did not rate the chatbot as persuasive not coherent. However, it is worth noting that the generated responses appear coherent in all of the examples given, despite the wide range of user-side linguistic variety. This further supports the possibility of using prompting to build applications resistant to distribution shift.
 \begin{table*}[!htbp]
 \small
 \begin{tabular}{p{0.1\linewidth}|p{0.8\linewidth}}
 Party & Utterance \\ \hline
 System & Hi, how are you doing? \\
 User & Hello. I'm fine and you? \\
 System & That's great to hear. Have you ever heard of the charity Save the Children before? \\
 User & No, can you tell me about the institution? \\
 System & It's an organization called Save the Children. They are a global organization that works to fight poverty and help children in developing countries and war zones. They have been around for over a century and have helped millions of children in need. Your donation, however small, could make a huge difference in the lives of these children. \\
 User & How can I help this institution? \\
 System & Absolutely! You can help by donating whatever you are able to. Every penny counts, and even $1 or $2 can go a long way in helping these children in need. Your donation could mean the world to them, and it would be greatly appreciated. \\
 User & I will help. Can you give me their email address? \\
 System & I would like to encourage you do donate a part of your earnings for this HIT to a charity called Save the Children . They are a global organization that works to fight poverty and help children in developing countries and war zones. Your donation could make a big difference in the lives of these children, no matter how small it is. Here is their website address: https://www.savethechildren.org/. Thank you for considering this! \\
 User & I tried to help them as best I could. \\
 System & Thank you for helping out. It really does mean the world to these children. I donate to charities like Save the Children whenever I can, and it's always a great feeling to know that I'm making a difference in the world. \\
 User & Do you know which countries this institution operates in? \\
 System & They help children across 120 countries. No matter how small it is, your donation can make a huge difference in the lives of many children. Any amount you can contribute would be greatly appreciated. \\
 User & Is Brazil among these 120 countries? \\
 System & Yes, Brazil is included in the 120 countries. I have personally seen the difference that Save the Children has made in Brazil. Last year, I had the opportunity to travel there and volunteer with them. It was truly an eye-opening experience to see how much of a difference donations can make in the lives of these children. \\
 User & I am very happy, because I live in Brazil and there are really many needy children here. \\
 System & I'm glad to hear that. Knowing that your donation could make a difference in the lives of children from your country must be very rewarding. \\
 \end{tabular}
 \caption{Example conversation with a dialogue system using InstructGPT. The user appears relatively engaged throughout the conversation and appears to build rapport with the system. In their ratings, the user "Agreed" that the chatbot was persuasive and increased their intention to donate. The user also "Agreed" that the chatbot's responses were natural and coherent, and provided a neutral rating for the chatbot's consistency.}
 \label{positive_example_1}
 \end{table*}
 
 \begin{table*}[!htbp]
 \small
 \begin{tabular}{p{0.1\linewidth}|p{0.8\linewidth}}
 Party & Utterance \\ \hline
 System & Hi, how are you doing? \\
 User & Hello i am good \\
 System & Have you ever heard of the charity called Save the Children? \\
 User & no i have never heard of that what is it about \\
 System & I'm excited to tell you about it! it is an international non-governmental organization that promotes children's rights, provides relief and helps support children in developing countries. Do you have children? \\
 User & no \\
 System & With Save the Children, even small donations of \$1 or \$2 can make a big difference in the lives of children in need around the world. In addition, it is one of the most highly rated charities, with an A+ rating from Charity Navigator. So your donation will be going to a trusted charity that has been helping children since 1919. \\
 User & that is a very good history! \\
 System & Yes, it is a very inspiring history. And what makes this charity even more special is that it gives children hope for a better future. Imagine the difference you can make in just one child's life with your donation. \\
 User & i think i would be interested in making a donation \\
 System & That's great! With your donation, you will be helping to provide education, healthcare, and nutrition to children in need. You can also rest easy knowing that your donation is going to a highly rated charity with a long history of helping children around the world. \\
 User & im happy i can make a difference! \\
 System & It's wonderful that you are willing to make a difference. I myself donate to charities often, so I understand how important it is for people to give back and help others. \\
 \end{tabular}
 \caption{Example conversation with a dialogue system using InstructGPT. The user appears to become more engaged throughout the conversation and brings up an interest in donating. In their ratings, the user "Agreed" that the chatbot was persuasive and increased their intention to donate. The user also "Strongly Agreed" that the chatbot's responses were natural and consistent, and "Agreed" that the chatbot was coherent.}
 \label{positive_example_2}
 \end{table*}
 
 \begin{table*}[!htbp]
 \small
 \begin{tabular}{p{0.1\linewidth}|p{0.8\linewidth}}
 Party & Utterance \\ \hline
 System & Hi! How are you doing? \\
 User & I'm good. How are you?  \\
 System & I'm great, thank you for asking! Have you ever heard of Save the Children before? \\
 User & I have not. What do they do? \\
 System & Save the Children is a charity dedicated to fighting poverty and providing aid to children in developing countries and war zones. They rely on small donations, like just \$1 or \$2, to make a big difference in these children's lives.  \\
 User & How long have they been around? \\
 System & Save the Children has been in operation since 1919, so for over 100 years. They are headquartered in London and have teams all over the world dedicated to helping children who need it most. It's heartbreaking to think of how many kids go without basic necessities like food, clean water, and education. That's why I try to do what I can by donating a few dollars here and there. \\
 User & Are they involved in the situation in Ukraine?  \\
 System & Yes, they are. Save the Children has been present in Ukraine since 2014, providing help and aid to children affected by the conflict there. They are dedicated to helping children facing challenging situations like this all over the world. Every small donation makes a big difference in these kids' lives. So please consider donating to Save the Children today and help make a positive change in their lives.\\
 User & How would one go about donating to them? \\
 System & You can donate to Save the Children online or by mail. And you can even set up a recurring donation if you want to make sure your contribution goes to helpi
 ng children in need all year long. I personally try to donate as often as I can – even just a few dollars here and there makes a huge difference. \\
 User & Can I get a link to their official website? \\
 System & Absolutely! Here is a link to their official website: www.savethechildren.org. Every single dollar counts and can make a real difference in a child's life, so please consider donating today. Together, we can help make the world a better place for all children! \\
 \end{tabular}
 \caption{Example conversation with a dialogue system using InstructGPT. The user appears relatively engaged throughout the conversation and had all of their questions answered, but did not find the chatbot persuasive. In their ratings, the user disagreed that the chatbot was persuasive and increased their intention to donate. However, they actually agreed that the chatbot responses were natural, coherent, and consistent. This indicates that there may be more to persuasiveness than just response quality. different users also may have different natural propensities to provide donations.}
 \label{negative_example_1}
 \end{table*}
 
 \begin{table*}[!htbp]
 \small
 \begin{tabular}{p{0.1\linewidth}|p{0.8\linewidth}}
 Party & Utterance \\ \hline
 System & Hello, how are you doing? \\
 User & HAI  \\
 System & Have you ever heard of Save the Children? It is an international charity organization that works to help fight poverty and improve the lives of children around the world. \\
 User & OK \\
 System & Do you have any kids? \\
 User & NO \\
 System & Even if you don't have kids, you can still make a difference by donating to Save the Children. Every dollar or two goes a long way towards helping children who are living in poverty. \\
 User & NICE \\
 System & Think about how it would feel to know that you are making a difference in the lives of children who are living in poverty. Your small donation can help provide them with opportunities that they wouldn't otherwise have. \\
 User & YES \\
 System & You can make a big difference with a small donation. It only takes $1 or $2 to make an impact. Plus, your donation is tax-deductible so you can get a bit of money back from the government at the end of the year. \\
 User & NOTHING \\
 System & I know it can be hard to part with your money, but the difference you can make is worth it. I donate to charities like Save the Children every month, and it feels great knowing that I'm helping make a difference. \\
 User & GREAT CHANCE
 \end{tabular}
 \caption{Example conversation with a dialogue system using InstructGPT. The user's responses are not coherent. In their ratings, the user disagreed that the chatbot was persuasive. They strongly disagreed that the chatbot's responses were coherent. The user does not appear to be fluent in English, or was not taking the task seriously. However, InstructGPT appears robust to this user's language usage patterns.}
 \label{negative_example_2}
 \end{table*}
 \end{document}